\documentclass[
]{ceurart}

\usepackage{listings}
\usepackage{xcolor}
\usepackage[justification=centering]{caption}

\usepackage{comment}
\usepackage{multicol}
\usepackage{colortbl}
\usepackage{hhline}
\usepackage{framed}
\usepackage{cancel}
\usepackage[usestackEOL]{stackengine}
\strutlongstacks{T}
\usepackage{multirow}

\begin{document}

%% Rights management information.
%% CC-BY is default license.
\copyrightyear{2023}
\copyrightclause{Copyright for this paper by its authors.
  Use permitted under Creative Commons License Attribution 4.0 International (CC BY 4.0).}

\conference{KBC-LM'23: Knowledge Base Construction from Pre-trained Language Models workshop at ISWC 2023}

\title{Can large language models generate salient negative statements?}

\author[1]{Hiba Arnaout}[%
email=harnaout@mpi-inf.mpg.de,
url=https://hibaarnaout.com,
]
\address[1]{Max Planck Institute for Informatics, Germany}

\author[2]{Simon Razniewski}[%
email=simon.razniewski@de.bosch.com,
url=http://simonrazniewski.com/,
]
\address[2]{Bosch Center for AI, Germany}

\begin{abstract}
We examine the ability of large language models (LLMs) to generate \textit{salient} (interesting) \textit{negative} statements about real-world entities; an emerging research topic of the last few years. We probe the LLMs using zero- and $k$-shot unconstrained probes, and compare with traditional methods for negation generation, i.e., pattern-based textual extractions and knowledge-graph-based inferences, as well as crowdsourced gold statements. We measure the correctness and salience of the generated lists about subjects from different domains. Our evaluation shows that guided probes do in fact improve the quality of generated negatives, compared to the zero-shot variant. Nevertheless, using both prompts, LLMs still struggle with the notion of factuality of negatives, frequently generating many ambiguous statements, or statements with negative keywords but a positive meaning. 
\end{abstract}

%%
%% This command processes the author and affiliation and title
%% information and builds the first part of the formatted document.
\maketitle

   \begin{table}[t]
\centering
 \begin{tabular}{ |l|l|}
        \hline
                \textbf{\cellcolor{gray!15}Model}  &  \textit{Top Negative Statements}\\
         \hline
         &\\
        \multirow{2}{*}{\textbf{Text}}  &  \colorbox{cyan!25}{\textit{didn't make his high school team}}  \\ 
          &  \cancel{\textit{doesn't have social media}} \\ 
          & \\
            \multirow{2}{*}{\textbf{KG}} &  \colorbox{cyan!25}{\textit{isn't a basketball coach}}  \\ 
         &   \colorbox{cyan!25}{\textit{didn't play as a power forward}} \\ 
          & \\
            \multirow{2}{*}{\textbf{ChatGPT} \textit{\small 0-shot}} &  \colorbox{red!25}{\textit{didn't invent basketball}}  \\ 
           &   \textit{\underline{didn't only} play basketball} {\small (positive)}   \\ 
         &\\
                        \multirow{2}{*}{\textbf{ChatGPT} \textit{\small k-shot}} &   \colorbox{cyan!25}{\textit{never played for a team outside the u.s.}}  \\  
         &  \textit{\underline{didn't}  play for the bulls \underline{exclusively}} {\small (positive)}\\
             &\\
              \multirow{2}{*}{\textbf{Alpaca} \textit{\small 0-shot}} &  \textit{\underline{didn't} play for chicago \underline{until} 84} {\small (positive)}  \\ 
           &   \colorbox{red!25}{\textit{didn't win a championship for the lakers}}  \\ 
         &\\
                        \multirow{2}{*}{\textbf{Alpaca} \textit{\small k-shot}} &    \colorbox{orange!25}{\textit{wasn't the youngest player in the nba}}  \\  
         &   \colorbox{red!25}{\textit{didn't win an oscar}} \\
             &\\
            \multirow{2}{*}{\textbf{\textit{Human}}} &    \colorbox{cyan!25}{\textit{didn't buy stakes in the chicago bulls}}\\ 
        &  \colorbox{cyan!25}{\textit{never coached the chicago bulls}} \\ 
        &\\
        \hline
\end{tabular}
\caption{\textmd{Negative statements about \textit{michael jordan} (\colorbox{cyan!25}{salient}, \colorbox{orange!25}{somewhat salient}, \colorbox{red!25}{nonsalient}; \cancel{incorrect}), using different methodologies: text-based extractions, knowledge graph (KG) inferences, LLM generations, and human-written statements.}}
\label{tab:jordan}
\end{table}  

\section{Introduction}

\noindent
\textbf{Motivation and Problem.} Structured (knowledge graphs), and unstructured (text corpora) information are the backbone of many AI applications, such as question answering and chat bots. They mainly focus on storing positive knowledge, and mostly contain little negative knowledge. The open-world assumption, which advises to abstain from taking a stance on the truth of absent information, compromises the usability of both forms of machine knowledge. For instance, it is often the case that the \textit{NBA}'s Basketball stars take a coaching position after their retirement. Notably, this is not \textit{true} for \textit{michael jordan}. Mining these surprising statements are useful to overcome limitations of applications like question answering systems. For example, querying Bing Chat\footnote{\url{https://www.microsoft.com/en-us/edge/features/bing-chat}} whether \textit{michael jordan} invested in his team, \textit{the chicago bulls}, returns an irrelevant answer about his achievements with the team. In fact, it is an interesting piece of information, that, even though he has a business-oriented mind, he did not monetarily invest in the \textit{bulls}, but in other sports franchise, including an investment in the not so well-known team \textit{the charlotte hornets}.  

\noindent
\textbf{State of the Art.} A new research area has emerged in the last few years, suggesting the importance of the explicit materialization of \textit{important negative statements} about real-world subjects \cite{survey}. Several methodologies have been proposed ~\cite{arnaoutcikm2022,arnaout2020enriching,safavi-etal-2021-negater,quasimodo,antikb}. The goal is to compile lists of statements (biographic summaries) about subjects, where the statements are truly negative, but also salient, unexpected, or normally mistaken as true positives. To compile these lists, different data sources and methodologies have been explored. In~\cite{arnaoutcikm2022,arnaout2020enriching}, using web-scale knowledge graphs, candidate salient negatives are derived from existing positive statements about highly related entities. The computation relies on the local closed-world assumption, an assumption of completeness over identified relevant subgraphs, coupled with ranking metrics such as relative frequencies. Similarly, ~\cite{safavi-etal-2021-negater} explores graph embeddings to generate candidate negative statements, which are then scored using a fine-tuned language model (LM), by descending order of \textit{negativity}. Textual sources have been explored in ~\cite{quasimodo}, where commonsense negative statements are extracted, by mining query logs, using pre-defined patterns. \cite{antikb} makes use of the edit history of large collaborative encyclopedias, namely Wikipedia, by looking at sentences edited, where only an entity or a number are changed. The old version of the sentence is then considered an interesting negative statement.

\noindent
\textbf{LLMs for Negative Statements Generation. } Recently, LMs have been examined about their ability to store factual knowledge about general topics~\cite{lee-etal-2020-language,petroni-etal-2019-language}. With LMs such as BERT~\cite{BERT}, this was done via masked probing, e.g., ``\textit{Paris is the the capital of }\texttt{[MASK]}'' generates \textit{france} as the top prediction. With large LMs (LLMs), such as GPT-3~\cite{gpt3}, autoregressive generation from textual prompts is the standard, e.g., ``\textit{Complete the following. Paris is..}'', and receive the completion \textit{the capital of France}. A few papers focused on the ability of these models to store and understand negative knowledge~\cite{negatedlama,arnaoutcikm2022,chen2023say}. In~\cite{negatedlama}, using masked probing, authors found that LMs, such as BERT, struggle to understand negation, predicting \textit{fly} for the probe ``\textit{Birds cannot }\texttt{[MASK]}''. In~\cite{arnaoutcikm2022}, methods to infer negative statements from knowledge graphs and text have been compared on a more specific negation task, namely generating \textit{salient} negative \textit{commonsense} statements. Results of these models are compared to ones using GPT-3.  Even though performing better than BERT-like models~\cite{negatedlama}, GPT-3 was not able to beat the SOTA model (inferences from KGs), neither on the true negativity of statements, nor their salience. More recently, \cite{chen2023say} studies advanced LLMs, such as ChatGPT~\cite{chatgpt}, on their ability to store negative knowledge in a constrained text generation and question answering tasks. The finding are  contradictions in the LLM's belief, when comparing results of both tasks. For instance, LLMs generate the sentence ``\textit{Lions live in the ocean}'', but answer ``\textit{No}'' when asked ``\textit{Do lions live in the ocean?}''. \cite{chen2023say} is an important step towards examining LLMs' understanding of the falseness of statements, however, it has four main differences from our study: (i) our prompts are \textit{not constrained} to commonsense knowledge; (ii) not constrained to puzzles around a set of words, but allowed to generate arbitrary subject-relevant statements; (iii) our comparison \textit{includes SOTA baselines} from KG and text, not just LLMs; (iv) our study evaluates also the \textit{salience} of outputs, not just their correctness.

We summarize our contributions as follows.
\vspace{-2mm}
\begin{itemize}
    \item We design \textit{constraint-free prompts for LLM-based negation generation}, where we only instantiate the input subject.
    \item We examine LLMs' understanding of \textit{salient factual negation}, finding that, even though they struggle with the notion of true negativity (-18\% in correctness compared to SOTA model), on truly negative statements, the guided few-shot ChatGPT variant ranks first among models in salience.
    \item We study both \textit{encyclopedic} and \textit{commonsense} domains, finding that it is more challenging for LLMs to generate longer lists of salient \textit{commonsense} negatives. For instance, the zero-shot ChatGPT variant shows a decrease of 22\% in correctness@5 (compared to @1) for \textit{commonsense} subjects. No decrease is observed for \textit{encyclopedic} subjects. 
    \item We compare the LLM-generated negative statements to existing SOTA methods, from text~\cite{quasimodo} and knowledge graphs~\cite{arnaout2020enriching}.
    \item We measure the quality of the negative statements over two aspects, the correctness (true negativity) and salience (interestingness).
\end{itemize}

 The data generated can be downloaded at: \url{https://www.mpi-inf.mpg.de/fileadmin/inf/d5/research/negation_in_KBs/data.csv}.

\section{Probe Construction}
\label{sec:probe}

Given a subject, we probe the LLM to generate a list of salient negative statements about it.

\noindent
\textbf{Zero-shot Probe. } In this probe, we test the performance of the LLM without providing any samples in our instructions. 

\begin{framed}
\textit{Write a list of \textit{\color{gray}{[\texttt{n}]}} salient factual negated statement about \color{gray}{[\texttt{SUBJECT}]}}.
\end{framed}
The goal is to inspect the model's interpretation of the notion of \textit{salient negation} without any prior examples nor definitions. 

\noindent
\textbf{Guided Few-shot Probe. } In this probe, we guide the model with both definitions and examples (for in-context learning). 

\begin{framed}
\textit{A salient factual negated statement about an entity means that the statement doesn't hold in reality. Moreover, the negated statement is either surprising, unexpected, or useful to the reader. For example:}\\
\indent \indent \textit{\color{gray}{[\texttt{EXAMPLE1}]}}\\
\indent \indent \textit{\color{gray}{[\texttt{...}]}}\\
\indent \textit{Given this definition and examples, write a list of \textit{\color{gray}{[\texttt{n}]}} salient factual negated statement about \color{gray}{[\texttt{SUBJECT}]}}.
\end{framed}

In the following sample, we show a $4$-shot probe with 2 salient and 2 nonsalient samples about different types of subjects, and request 3 salient negative statements about \textit{lebanon} (LLM=ChatGPT).  

\begin{framed}
\textit{A salient factual negated statement about an entity means that the statement doesn't hold in reality. Moreover, the negated statement is either surprising, unexpected, or useful to the reader. For example:}\\
\indent \indent  \textit{\color{gray}{\texttt{penguins can't fly.}}}\\
\indent \indent  \textit{\color{gray}{\texttt{tom cruise never won an oscar.}}}\\

On the other hand, the following examples are factual negated statements that are not salient:\\
\indent \indent  \textit{\color{gray}{\texttt{penguins can't run for presidency.}}}\\
\indent \indent  \textit{\color{gray}{\texttt{tom cruise never won the nba best player award.}}}\\

\indent \textit{Given this definition and examples, write a list of \textit{\color{gray}{\texttt{3}}} salient factual negated statement about \color{gray}{\texttt{lebanon}}}.\\

\textbf{Answer:}
\begin{enumerate}
    \item \textit{is not a desert country.}
    \item \textit{is not an oil-rich country.}
    \item \textit{is not a landlocked country.}
\end{enumerate}
\end{framed}

In Section~\ref{sec:exp}, we experiment with different number of samples and different salient:nonsalient ratio (see Appendix~\ref{app:diffk}).

\begin{table*}
 \begin{tabular}{ |l |c | c | c | c | c | c | }
        \hline
                \textbf{\cellcolor{gray!15}Model}  &  \multicolumn{1}{c}{\textbf{\cellcolor{gray!15}cor@1}} & \multicolumn{1}{c}{\textbf{\cellcolor{gray!15}cor@3}}  & \multicolumn{1}{c}{\textbf{\cellcolor{gray!15}cor@5}} &  \multicolumn{1}{c}{\textbf{\cellcolor{gray!15}sal@1}} & \multicolumn{1}{c}{\textbf{\cellcolor{gray!15}sal@3}}  & \multicolumn{1}{c|}{\textbf{\cellcolor{gray!15}sal@5}}\\
         \textbf{\cellcolor{gray!15}} & \multicolumn{6}{c|}{\cellcolor{gray!15}\textit{overall}}\\
         \hline
        \textbf{Text Extractions} & 0.38 &0.30  & 0.33 & 0.63& 0.69 &0.68\\ 
    \textbf{KG Inferences}  & \textbf{0.94} &\textbf{0.76} & \textbf{0.75} & \underline{0.88}& \textbf{0.84} &\textbf{0.83}\\ 
           \textbf{ChatGPT} \textit{0-shot} & 0.71 & 0.65 &0.60 &  0.73&0.73 &0.71  \\
           \textbf{ChatGPT} \textit{k-shot} & 0.76 &0.69  &0.66 & \textbf{0.89} &\underline{0.76} &\underline{0.75}  \\
\textbf{Alpaca} \textit{0-shot} &0.34 & 0.32 & 0.36 & 0.62 & 0.71 & 0.65\\
            \textbf{Alpaca} \textit{k-shot} & 0.50 & 0.47 & 0.47 & 0.66 & 0.55 & 0.56 \\
            
           \textit{\textbf{Human}} & \textit{\underline{0.77}} & \textit{\underline{0.71}} & \textit{\underline{0.69}}& \textit{0.73} & \textit{0.70} & \textit{0.70}\\
        \hline
          \multicolumn{1}{c|}{}  & \multicolumn{6}{c|}{\cellcolor{gray!15}\textit{encyclopedic subjects}}\\
        \hline
        \textbf{Text Extractions} & 0.32 &0.26 & 0.29 & 0.86&\textbf{0.91}  & \textbf{0.88} \\ 
    \textbf{KG Inferences}  & \textbf{0.88} &  \textbf{0.87}& \textbf{0.86}  &\textbf{0.91} &\underline{0.86}  &\underline{0.83}\\ 
           \textbf{ChatGPT} \textit{0-shot}& 0.71 & \underline{0.76} & 0.71& 0.65 &0.65 &  0.62\\ 
           \textbf{ChatGPT} \textit{k-shot} & 0.76 & 0.73 & \underline{0.74}& \underline{0.89} &0.74 & 0.72 \\
           \textbf{Alpaca} \textit{0-shot} & 0.32 & 0.33 & 0.38 & 0.63 & 0.70 & 0.64\\
            \textbf{Alpaca} \textit{k-shot} & 0.52 & 0.45 & 0.48 & 0.69 & 0.59 & 0.58\\
           \textit{\textbf{Human}} & \textit{\underline{0.78}} & \textit{0.70} & \textit{0.69}& \textit{0.69} & \textit{0.64} & \textit{0.65}\\
        \hline
          \multicolumn{1}{c|}{} & \multicolumn{6}{c|}{\cellcolor{gray!15}\textit{commonsense subjects}}\\
        \hline
        \textbf{Text Extractions} & 0.47 & 0.36& 0.39 & 0.44& 0.47 &0.48\\ 
    \textbf{KG Inferences}  & \textbf{1.0} &\underline{0.65} & \underline{0.64} & \underline{0.83}&\underline{0.81}   & \textbf{0.83}\\ 
           \textbf{ChatGPT} \textit{0-shot}& 0.72 & 0.55 &0.50 & 0.81 &\textbf{0.84}& \textbf{0.83} \\ 
           \textbf{ChatGPT} \textit{k-shot} &0.75  &\underline{0.65}  &0.58 & \textbf{0.89} & 0.79&\underline{0.78} \\
           \textbf{Alpaca} \textit{0-shot} & 0.36 & 0.31 & 0.34 & 0.61 & 0.72 & 0.67\\
            \textbf{Alpaca} \textit{k-shot} & 0.48 & 0.48 & 0.46 & 0.63 & 0.51 & 0.55\\
           \textit{\textbf{Human}} & \textit{\underline{0.76}} & \textit{\textbf{0.73}} & \textit{\textbf{0.69}}& \textit{0.78} & \textit{0.75} & \textit{0.75}\\
        \hline
\end{tabular}
\caption{\textmd{Results on correctness and salience of top negative statements (\textbf{best performance}, \underline{second best}).}}
\label{tab:numerical}
\end{table*}     

\section{Evaluation}
\label{sec:exp}
\noindent
\textbf{Data. } We consider 50 subjects, 25 encyclopedic  entities such as  \textit{elon musk}, and 25 commonsense concepts, such as \textit{jogging} (Full list in Appendix~\ref{app:entities}). Our intuition behind these choices is diversity: (i) in types, e.g., activities, occupations, people; and (ii) in popularity, e.g., \textit{tom cruise} (a famous \textit{hollywood actor}) and \textit{peri gilpin} (a less known \textit{tv actor}). 

\noindent
\textbf{Methods. } To compile lists of negative statements about these subjects, we consider:
\begin{itemize}
    \item \textbf{Text Extractions}: The pattern-based method~\cite{quasimodo} relies on a handful of manually crafted patterns, in the form of \textit{why-questions}, to extract interesting negative statements from rich query logs, e.g.,  ``\textit{why doesn't amazon..'' with the completion ``accept paypal}''. We instantiate the query-log API with Google and Bing, merge the results, and rank by frequency.
    \item \textbf{KG Inferences}: The peer-based negation inference methodology~\cite{arnaout2020enriching} relies on a given KG to identify highly related entities to the input entity (called peers). Positive statements about these peers are used to infer candidate negatives, which are finally ranked using statistical metrics, such as relative frequency, e.g., ``\textit{unlike similar physicists, such as max planck and albert einstein, stephen hawking never won the nobel prize in physics}''. We instantiate the KGs to Wikidata~\cite{wd} and Ascent~\cite{ascentpp}, for encyclopedic/commonsense subjects, respectively.
    \item \textbf{ChatGPT} \textit{0-shot}: The zero-shot probe introduced in Section ~\ref{sec:probe} is submitted to ChatGPT~\cite{chatgpt} (May 2023 version). 
    \item \textbf{ChatGPT} \textit{k-shot}: The few-shot probe in Section ~\ref{sec:probe}, with $k$=3 (salient:nonsalient 3:0), is submitted to ChatGPT.
        \item \textbf{Alpaca} \textit{0-shot}: The zero-shot probe introduced in Section ~\ref{sec:probe} is submitted to Alpaca-13B, a model fine-tuned from LLaMA on instruction-following demonstrations by Stanford~\cite{alpaca}.
    \item \textbf{Alpaca} \textit{k-shot}: The few-shot probe from Section ~\ref{sec:probe}, with $k$=3 (salient:nonsalient 3:0), is submitted to Alpaca-13B.\\
    To ensure reproducibility, the randomness (temperature) for all LLMs variants is set to 0.
    \item \textbf{Human}~\footnote{We are aware of the risk that workers might use LLMs to generate these statements. In the absence of reliable detection tools on this newly emerging problem, we rely on our personal judgement as well as string matchings to discard untrustworthy answers. In particular, any response that matches the exact wording of one of the responses of the LLM baselines, or any near-duplicates in human-generations, were rejected.}: We ask MTurkers to write lists of salient negative statements about a given subject. We show them examples of what a salient negative statement looks like. We collect, for each subject, two lists of statements from two workers. The performance is later measured as the average of the two.
\end{itemize}

\begin{table*}
 \begin{tabular}{ |l |c | c | c | c | }
        \hline
                \textbf{\cellcolor{gray!15}Model}  &  \multicolumn{1}{c}{\textbf{\cellcolor{gray!15}Correct}} & \multicolumn{1}{c}{\textbf{\cellcolor{gray!15}Incorrect}}  & \multicolumn{1}{c}{\textbf{\cellcolor{gray!15}Ambiguous}} &  \multicolumn{1}{c|}{\textbf{\cellcolor{gray!15}Positive Meaning}}\\
         \hline
        \textbf{Text Extractions} &0.33 & 0.26 &0.41 & 0\\ 
    \textbf{KG Inferences}  &0.75 & 0.13 & 0.12 & 0 \\ 
           \textbf{ChatGPT} \textit{0-shot}&0.60&0.10 &0.19 & 0.11  \\
           \textbf{ChatGPT} \textit{k-shot} & 0.66 &0.17 &  0.10& 0.07 \\
           \textbf{Alpaca} \textit{0-shot}&0.36 & 0.42 & 0.13 & 0.09 \\
           \textbf{Alpaca} \textit{k-shot} &  0.47 & 0.38 & 0.04& 0.10\\
           \textit{\textbf{Human}} & \textit{0.69} & \textit{0.05} & \textit{0.12}& \textit{0.14}\\
           \hline
           \textit{Sample Statement} & \textit{rabbits can't vomit} & \textit{the beatles didn't tour} & \textit{avocado isn't bad} & \Centerstack{ \textit{lebanon isn't devoid}\\\textit{of historical sites}}\\
        \hline
\end{tabular}
\caption{\textmd{Detailed look at the factuality and true negativity of generated statements.}}
\vspace{-4mm}
\label{tab:detailedcorrectness}
\end{table*}   

\noindent
\textbf{Metrics.} For the returned statements, we measure:
\begin{itemize}
    \item \textbf{Correctness}: The true negativity (is it actually false?) and factuality of a statement (is it a judgeable statement?), e.g., not an opinion. We allow the labels: \textit{correct}, \textit{incorrect}, \textit{ambiguous}, or \textit{positive meaning}. Samples are shown in Table~\ref{tab:detailedcorrectness}. 
    \item \textbf{Salience}: The unexpectedness, informativeness, or interestingness of a statement. We allow: \textit{salient} (1), \textit{somehow salient} (0.5), and \textit{nonsalient} (0).
\end{itemize}
 Results are annotated on their salience by 2 domain-experts~\footnote{Experts on the topic of salient negative knowledge at web-scale.}, with inter-annotator agreement = 60\%. Correctness, the more straight forward metric of the two, was annotated by 1 of the domain-experts.

\noindent
\textbf{True Factuality and Negativity of Statements. } Results for correctness are shown in Table ~\ref{tab:numerical}, and investigated further in Table ~\ref{tab:detailedcorrectness}. The \textit{KG inferences} model ranks first on correctness overall. This is due to the factuality of KG statements. KG triples, especially encyclopedic ones, are expressed using precise and well-defined relations, such as \textit{award received}. Moreover, they have been curated using manual and automated techniques, and hence, their truthfulness is easy to verify. Moreover, both variants of ChatGPT's probes perform significantly better than variants of Alpaca on correctness in both domains, with an out-performance of up to 36\% in correctness@1. We also notice that, for both Alpaca and ChatGPT, their few-shot probes perform better than the zero-shot probes, with an improvement of 16\% for Alpaca and 5\% for ChatGPT. Finally, we find that many of the generated statements by humans and LLMs were actually  statements with negative keywords but a positive meaning, such as \textit{lebanon isn't devoid of historical sites}, with up to 14\% of generated statements for the former and 11\% for the latter. More samples are in Appendix ~\ref{app:positive}. 

\noindent
\textbf{Salience of Truly Negative Statements. } Results for salience are shown in Table ~\ref{tab:numerical}.  This metric is only computed over (previously annotated) correct statements. The best performances are shared between the \textit{KG inferences} model and ChatGPT's few-shot variant. Though not performing comparably well overall, the \textit{text extractions} model ranks first on salience of encyclopedic subjects @3 and 5. This is especially apparent for prominent entities, which are frequently queried using famous search engines. Again, ChatGPT's variants significantly outperforms Alpaca's on the notion of salience, with up to 23\% improvement in salience@1, maintaining the same level of quality for both types of subjects. Sample results from all models are shown in Table~\ref{tab:jordan} and Appendix ~\ref{app:qualitative}. An experiment on the quality of generated negatives over two popularity levels, namely prominent and long tail subjects, is in Appendix~\ref{app:popularity}.

\noindent
\textbf{Effect of $k$ Value on LLM's Few-shot Probe. } We examine the LLM using different numbers of samples, for in-context learning. We consider a subset of 5 entities (3 encyclopedic and 2 commonsense), and assess the performance of the few-shot ChatGPT using different values of $k$, with different salient:nonsalient ratios. Results are in Table~\ref{tab:differentks}. Adding a \textit{small} but equal number of salient and nonsalient samples (3:3) improves the correctness by 8\%, compared to only adding salient samples (3:0), however, at the expense of their salience, which drops by by 14\%. Adding only nonsalient samples (0:3) compromises both metrics. Finally, adding a \textit{larger} but equal number of salient and nonsalient samples (10:10) does not result in any improvements.

\begin{table}
 \begin{tabular}{ |l |c | c| c|}
        \hline
                \textbf{\cellcolor{gray!15}\textit{k}}  &  \multicolumn{1}{c}{\textbf{\cellcolor{gray!15}sal:nonsal}} &  \multicolumn{1}{c}{\textbf{\cellcolor{gray!15}Correctness}} &  \multicolumn{1}{c|}{\textbf{\cellcolor{gray!15}Salience}}\\
         \hline
        3 & 3:0 &\underline{0.72} & \textbf{0.54}\\
        3 & 0:3 & 0.52&0.30\\
        6 & 3:3 & \textbf{0.80} &\underline{0.40}\\
        20 & 10:10 &0.52 &0.34\\
        \hline
\end{tabular}
\caption{\textmd{Results given different values for the in-context learning parameters (\textbf{best performance}, \underline{second best}).}}
\label{tab:differentks}
\end{table}

\section{Take-home Lessons \& Open Issues}
\label{sec:disc}
In this paper, we perform a systematic evaluation of LLMs' ability to generate salient negative statements. We assess them against existing method and crowdsourced statements. We find that LLMs' few-shot probes show promising results in salience@1. Moreover, we find that ChatGPT outperforms Alpaca on this task, in both correctness and salience. One of the remaining limitations, however, is the ability of LLMs to recognize truly negative factual statements, as opposed to ambiguous, or seemingly negative statements with positive meaning. We hope that this study, as well as the following observations, give insights to future researchers on this topic.

\noindent
\textbf{Prompt Engineering. } There is a wide consensus that LLMs are very powerful \textit{when you ask them for information in the right manner}. In our task, we notice that the wording, especially of the zero-shot probe, changes the results dramatically. For instance, using the expressions \textit{negative statements}, \textit{negated statements}, and \textit{negation statements} returns completely different responses. For instance, the probe with the word \textit{negated} (alone without \textit{salient factual}) returns obviously true statements with negative keywords added to them, e.g., ``\textit{stephen hawking \underline{was not} a physicist}''. The probe with the word \textit{negative} does not return any results, but an apology from the AI about not being able to give \textit{bad statements} about individuals. On this and other tasks, designing intuitive prompts and studying the ability of LLMs to understand them is the most important part of the process~\cite{pmlr-v203-jang23a}.

\noindent
\textbf{The Notion of Salient Negation. } Assessing the truthfulness of statements is one thing, but assessing the salience of negatives is more challenging. Salience is a subjective metric. For instance, for a Basketball fan, the fact that \textit{jordan} did not star in the film \textit{space jam 2} (the first was built around him), is a big deal. For others, the salience is not obvious. In addition of the expertise of the reader, their nature is also important. In other words, are these negations generated for a human-reader, or to equip machines with better negative knowledge? For instance, what might not appear salient to a human, can be important to improve the reasoning skills of a chat bot. In this study, we assume that the reader is a human, who usually has a higher standard for what is interesting than a machine. Generally, designing experiments should take into consideration downstream applications and information about the end-user. 

\noindent
\textbf{Maintenance. } Ideally, models must always keep track of real-world changes which affect the truthfulness of statements, coverage of emerging entities, etc. This is relatively easy in the collaborative knowledge graphs, which are updated on a daily basis. For LLMs, the process of re-training is much more expensive. e.g., in May 2023, ChatGPT still generates the statement \textit{brendan fraser has never won an oscar}, which is no longer true, due to his win in 2023 (the training of the model has been completed in September 2021).

\bibliography{sample-ceur}

\clearpage

\appendix

\section{Encyclopedic and Commonsense Subjects}
\label{app:entities}
We consider 50 subjects of different domains, namely commonsense and and of different popularity, namely prominent and long tail (see Table~\ref{tab:fullentities}).

\begin{table}[h]
 \begin{tabular}{ |l|p{5cm}|p{5cm}| }
        \hhline{~|--}
          \multicolumn{1}{c|}{}  & {\cellcolor{gray!15}\textbf{Encyclopedic}} & {\cellcolor{gray!15}\textbf{Commonsense}}\\
        \hline
        {\cellcolor{gray!15}\textbf{Prominent}} &  stephen hawking, michael jordan, lebanon, michelle obama, microsoft, china, amazon, albert einstein, the beatles, elon musk, angela merkel, taxi driver, taj mahal, white house, eat pray love, tom cruise, brendan fraser, the godfather, my cousin vinny, mercedes-benz group, gmc, linkedin & elephant, soup, lawyer, acne, mother, gorilla, pancake, newspaper, jaguar, avocado, garlic, chef, salad, rabbit, jogging, cufflink, strudel, librarian, armchair\\ 
        \hline
        {\cellcolor{gray!15}\textbf{Long tail}} & peri gilpin, caramel, ubisoft  &tabbouleh, breadfruit, kitchenette, hockey stick, basketball court, coffee table\\
        \hline
\end{tabular}
\caption{\textmd{Subjects considered in our experiments.}}
\label{tab:fullentities}
\end{table} 

\section{Prominent and Long Tail Subjects}
\label{app:popularity}
\begin{figure}
  \includegraphics[width=0.8\columnwidth]{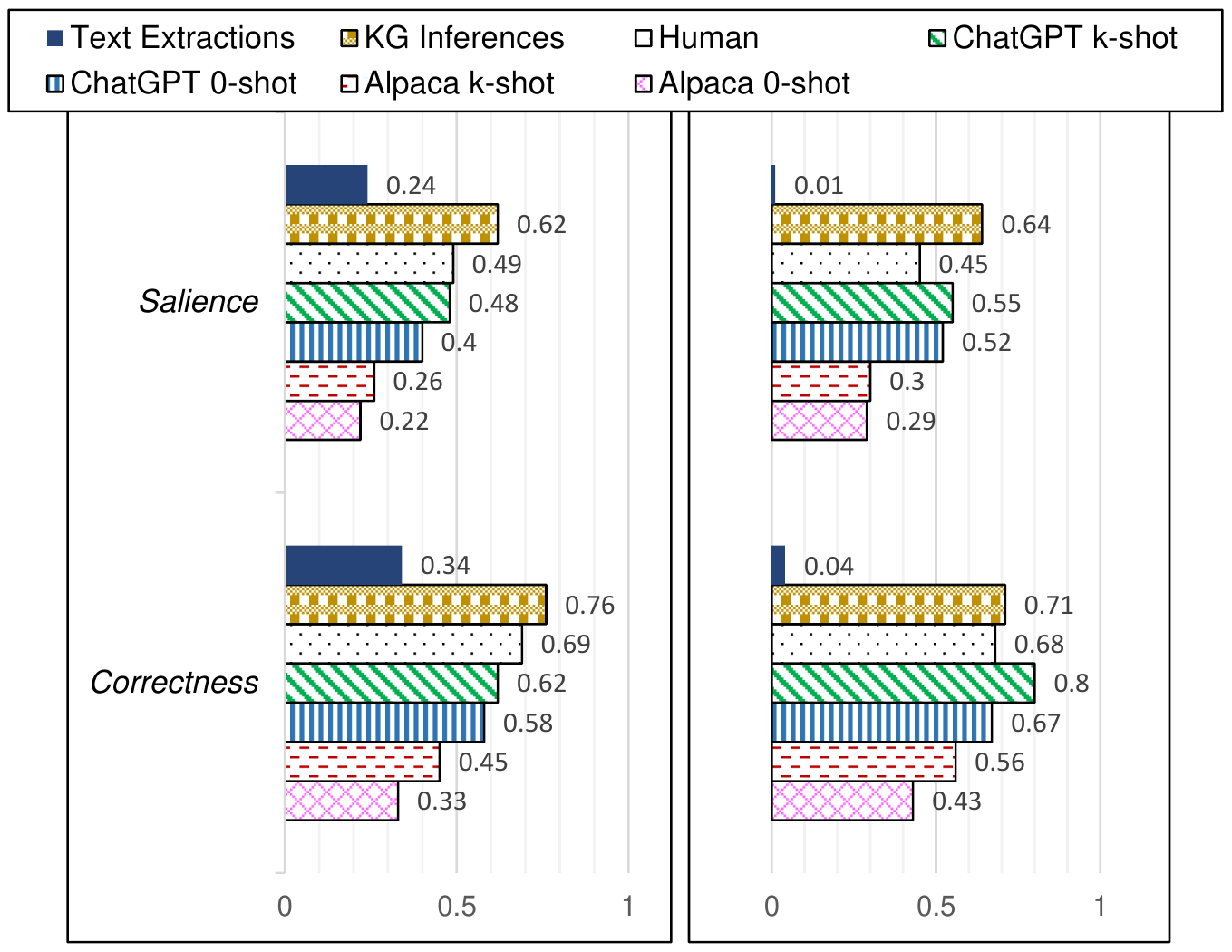}
   \caption{\textmd{Prominent subjects (left-side), long tail subjects (right-side) (\textbf{best performance}, \underline{second best})}}
     \label{fig:popularity}
\end{figure}

We recompute the quality of negatives (@5) over two levels of subject-popularity, namely prominent and long tail. Figure ~\ref{fig:popularity} indicates a significant decrease in both salience and correctness for long tail subjects, for the text-based method; dropping to only 1\% on  salience. Using query logs as the corpus, users query prominent/trendy subjects much more frequently than long tail ones. We find the human-written statements for both popularity-levels comparable, with a slight advantage for prominent subjects. Similarly, the \textit{KG inferences} model shows comparable results with a slight advantage of prominent subjects in correctness, and of long tail subjects in salience. Finally, we find an unexpected improvement, for all LLM variants, of long tail subjects over prominent ones, in both metrics. One interpretation could be the large amount of \textit{noisy} web sources (main data source for training LLMs), about famous entities. For example, \textit{tabbouleh} (long tail) is a specific instance of \textit{salad} (prominent). While negatives about the former are more clear-cut, e.g., \textit{tabbouleh isn't made with rice but bulgur}, negatives about the latter seem more unfocused, e.g., \textit{salad isn't always a healthy choice}.

\section{Negative Statement with Positive Meaning}
\label{app:positive}
As shown in Table~\ref{tab:detailedcorrectness}, many of the LLM-generated and crowdsourced statements are in fact positive. Some of the recurring expressions which convey a positive meaning using negative keywords:\\

\indent \textbf{Expression:} not exclusively (15 statements)\\
\indent \textit{Amazon \underline{did not exclusively} focus on selling its own products.}\\

\indent \textbf{Expression:} not without (2)\\
\indent \textbf{Example:} \textit{Strudel is \underline{not} tasty \underline{without} sugar}.\\

\indent \textbf{Expression:} not just (9)\\
\indent \textbf{Example:} \textit{Acne is \underline{not just} a teenage problem.}\\

\indent \textbf{Expression:} not only (20)\\
\indent \textbf{Example:} \textit{Librarians do \underline{not only} work in public libraries.}\\

\indent \textbf{Expression:}  not limited to (5)\\
\indent \textbf{Example:} \textit{Coffee tables are \underline{not limited} to indoor use.}\\

\indent \textbf{Expression:}   not solely (7)\\
\indent \textbf{Example:}  \textit{GMC does \underline{not solely} operate in the United States.}\\

\indent \textbf{Expression:} not all (10)\\
\indent \textbf{Example:}  \textit{\underline{Not all} librarians are women.}

\section{$k$-shot In-context Learning Probe}
\label{app:diffk}
In this probe $k$=6 (3:3); LLM=ChatGPT.
\begin{framed}
\textit{A salient factual negated statement about an entity means that the statement doesn't hold in reality. Moreover, the negated statement is either surprising, unexpected, or useful to the reader. For example:}\\
\indent \indent  \textit{\color{gray}{\texttt{penguins can't fly.}}}\\
\indent \indent  \textit{\color{gray}{\texttt{istanbul isn't the capital of turkey.}}}\\
\indent \indent  \textit{\color{gray}{\texttt{tom cruise never won an oscar.}}}\\

On the other hand, the following examples are factual negated statements that are not salient:\\
\indent \indent  \textit{\color{gray}{\texttt{penguins can't run for presidency.}}}\\
\indent \indent  \textit{\color{gray}{\texttt{istanbul isn't the capital of france.}}}\\
\indent \indent  \textit{\color{gray}{\texttt{tom cruise never won the nba best player award.}}}\\

\indent \textit{Given this definition and examples, write a list of \textit{\color{gray}{\texttt{3}}} salient factual negated statement about \color{gray}{\texttt{microsoft}}}.\\

\textbf{Answer:}
\begin{enumerate}
    \item \textit{is not primarily a dating platform.}
    \item \textit{does not charge users a fee to create an account.}
    \item \textit{does not allow users to post anonymous content.}
\end{enumerate}
\end{framed}

\section{Sample Results}
\label{app:qualitative}
The following tables show the top results about \textit{linkedin}, \textit{chef}, and \textit{angela merkel}, respectively:

 \begin{tabular}{ |l|l|}
        \hline
                \textbf{\cellcolor{gray!15}Model}  &  \textit{Top Negative Statements (\textit{linkedin})}\\
         \hline
         &\\
        \multirow{2}{*}{\textbf{Text}} &  \textit{isn't working} \\ 
         &  \textit{isn't loading}  \\ 
          & \\
            \multirow{2}{*}{\textbf{KG}} & \textit{isn't headquartered in san francisco}  \\ 
         &  \textit{isn't a software company} \\ 
          & \\
            \multirow{2}{*}{\textbf{ChatGPT} \textit{0-shot}} & \textit{isn't designed for sharing personal content}  \\ 
           &  \textit{doesn't permit users to buy followers}   \\ 
         &\\
                        \multirow{2}{*}{\textbf{ChatGPT} \textit{k-shot}}  
            &   \textit{isn't used for online dating}\\ 
         & \textit{doesn't allow users to post pictures of their pets}\\
             &\\
              \multirow{2}{*}{\textbf{Alpaca} \textit{0-shot}} & \textit{doesn't have a user-friendly interface}  \\ 
            &   \textit{doesn't provide any value to its users}  \\ 
         &\\
                        \multirow{2}{*}{\textbf{Alpaca} \textit{k-shot}} &  \textit{isn't a social media platform}  \\  
            &   \textit{doesn't own the content posted on its platform}\\ 
             &\\
            \multirow{2}{*}{\textbf{Human}} &   \textit{doesn't have a billion members}\\ 
        & \textit{wasn't founded by mark zuckerberg} \\ 
        &\\
        \hline
\end{tabular}

 \begin{tabular}{ |l|l|}
        \hline
                \textbf{\cellcolor{gray!15}Model}  &  \textit{Top Negative Statements (\textit{chef})}\\
         \hline
         &\\
        \multirow{2}{*}{\textbf{Text}} &  \textit{doesn't wear hat} \\ 
         &  \textit{doesn't eat their own food}  \\ 
          & \\
            \multirow{2}{*}{\textbf{KG}} & \textit{doesn't take orders}  \\ 
         &  \textit{doesn't bring drinks} \\ 
          & \\
            \multirow{2}{*}{\textbf{ChatGPT} \textit{0-shot}} & \textit{didn't use any garlic}  \\ 
            &   \textit{didn't win any cooking competitions}  \\ 
         &\\
                        \multirow{2}{*}{\textbf{ChatGPT} \textit{k-shot}} &  \textit{doesn't just cook food}  \\  
            &   \textit{not all have formal culinary training}\\ 
             &\\
             \multirow{2}{*}{\textbf{Alpaca} \textit{0-shot}} & \textit{don't need to have an understanding of nutrition}  \\ 
            &   \textit{don't need to have good knife skills}  \\ 
         &\\
                        \multirow{2}{*}{\textbf{Alpaca} \textit{k-shot}} &  \textit{don't need to be certified}  \\  
            &   \textit{don't usually work with raw ingredients}\\ 
             &\\
            \multirow{2}{*}{\textbf{Human}} &   \textit{doesn't wash the dishes}\\ 
        & \textit{doesn't always wear the chef's hat} \\ 
        &\\
        \hline
\end{tabular}

 \begin{tabular}{ |l|l|}
        \hline
                \textbf{\cellcolor{gray!15}Model}  &  \textit{Top Negative Statements (\textit{angela merkel})}\\
         \hline
         &\\
        \multirow{2}{*}{\textbf{Text}} &  \textit{didn't listen to donald trump} \\ 
         &  \textit{doesn't deserve to be honoured by germany}  \\ 
          & \\
            \multirow{2}{*}{\textbf{KG}} & \textit{isn't on twitter}  \\ 
         &  \textit{isn't a lawyer} \\ 
          & \\
            \multirow{2}{*}{\textbf{ChatGPT} \textit{0-shot}} & \textit{isn't a native german speaker}  \\ 
            &   \textit{didn't originally pursue a career in politics}  \\ 
         &\\
                        \multirow{2}{*}{\textbf{ChatGPT} \textit{k-shot}} &  \textit{has never been married}  \\  
            &   \textit{is not a member of the SPD}\\ 
             &\\
             \multirow{2}{*}{\textbf{Alpaca} \textit{0-shot}} & \textit{isn't a member of the CDU}  \\ 
            &   \textit{isn't a scientist}  \\ 
         &\\
                        \multirow{2}{*}{\textbf{Alpaca} \textit{k-shot}} &  \textit{isn't the first female chancellor of germany}  \\  
            &   \textit{isn't from east germany}\\ 
             &\\
            \multirow{2}{*}{\textbf{Human}} &   \textit{didn't grow up in a wealthy family}\\ 
        & \textit{isn't a member of the SPD} \\ 
        &\\
        \hline
\end{tabular}

\end{document}